\pdfoutput=1
\documentclass[10pt,twocolumn,letterpaper]{article}

\usepackage{3dv}
\usepackage{times}

\usepackage{amsmath}
\usepackage{amssymb}
\usepackage{balance}
\usepackage{bbm}
\usepackage{booktabs}
\usepackage{dsfont}
\usepackage{epsfig}
\usepackage{pifont}
\usepackage{graphicx}
\usepackage{grffile} 
\usepackage{multirow}
\usepackage[numbers,sort,compress]{natbib}
\usepackage{subcaption}
\usepackage{times}
\usepackage{url}
\usepackage[bottom]{footmisc}
\usepackage{comment}
\usepackage[table]{xcolor}

\newcommand{\cmark}{\ding{51}}%
\newcommand{\xmark}{\ding{55}}
\definecolor{aliceblue}{rgb}{0.94, 0.97, 1.0}

\def\sepappendix{0}

\usepackage[colorlinks,breaklinks=true,bookmarks=false]{hyperref}

\threedvfinalcopy %

\def\threedvPaperID{71} %

\begin{document}

\title{Towards Unconstrained Joint Hand-Object Reconstruction From RGB Videos}

\author{
 Yana Hasson\textsuperscript{1,2}
 \qquad G\"{u}l Varol\textsuperscript{3}
 \qquad Cordelia Schmid\textsuperscript{1,2}
 \qquad Ivan Laptev\textsuperscript{1,2}
 \\ \\
 {\normalsize \textsuperscript{1}Inria, \textsuperscript{2}D\'{e}partement d'informatique de l'ENS, CNRS, PSL Research University} \\
 {\normalsize \textsuperscript{3} LIGM,  $\Acute{\textrm{E}}$cole des Ponts, Univ Gustave Eiffel, CNRS, France} \\
{\normalsize \url{https://hassony2.github.io/homan.html}}
}

\maketitle

\begin{abstract}

Our work aims to obtain 3D reconstruction of hands and manipulated objects from monocular videos. 
Reconstructing hand-object manipulations holds a great potential for robotics and learning from human demonstrations.
The supervised  learning  approach  to  this  problem,  however,  requires 3D supervision and remains limited to constrained laboratory  settings  and  simulators  for  which  3D  ground truth is available.
In this paper we first propose a learning-free fitting approach for hand-object reconstruction  which can seamlessly handle two-hand object interactions.
Our method relies  on  cues  obtained  with  common  methods  for  object detection, hand pose estimation and instance segmentation. 
We quantitatively evaluate our approach and show that it can be applied to datasets with varying levels of difficulty for which training data is unavailable.

\end{abstract}
\section{Introduction}
\label{sec:intro}

Joint reconstruction of hand-object interactions
in the context of object manipulation has a broad range of practical
applications in robotics and augmented reality.
Yet, reliable and generic models for recovering 3D hand-object
configurations from images do not exist to date.
Learning-based approaches typically do not generalize
beyond their specific training domains.
Furthermore,
obtaining precise 3D annotations for supervision
is tedious and requires additional sensors.
Existing datasets rely on depth data~\cite{Hampali_2020_CVPR,chao:cvpr2021},
visible marker equipment~\cite{fphab_hernando_2018_cvpr,Pham_2018_TPAMI} or multiple views~\cite{interhand26m,Zimmermann_2019_ICCV,Simon_2017_CVPR,contactpose_Brahmbhatt_2020_ECCV}.
Moreover, such datasets typically feature a limited number of
objects~\cite{Tzionas:ICCV:2015,Tzionas:IJCV:2016,Mueller2017ICCV,RealtimeHO_ECCV2016} while still 
requiring manual annotations~\cite{chao:cvpr2021} or post-processing~\cite{Zimmermann_2019_ICCV,Simon_2017_CVPR}.
Synthetic rendering techniques can be used to create labeled datasets
with a larger number of objects, however, the resulting images lack realism in terms of appearance and grasp configurations~\cite{hasson19_obman,Zimmermann_2017_ICCV,GANeratedHands_CVPR2018}.

\begin{figure}[t]
    \begin{center}
        \includegraphics[width=0.9\linewidth,bb=0cm 0cm 18cm 18cm]{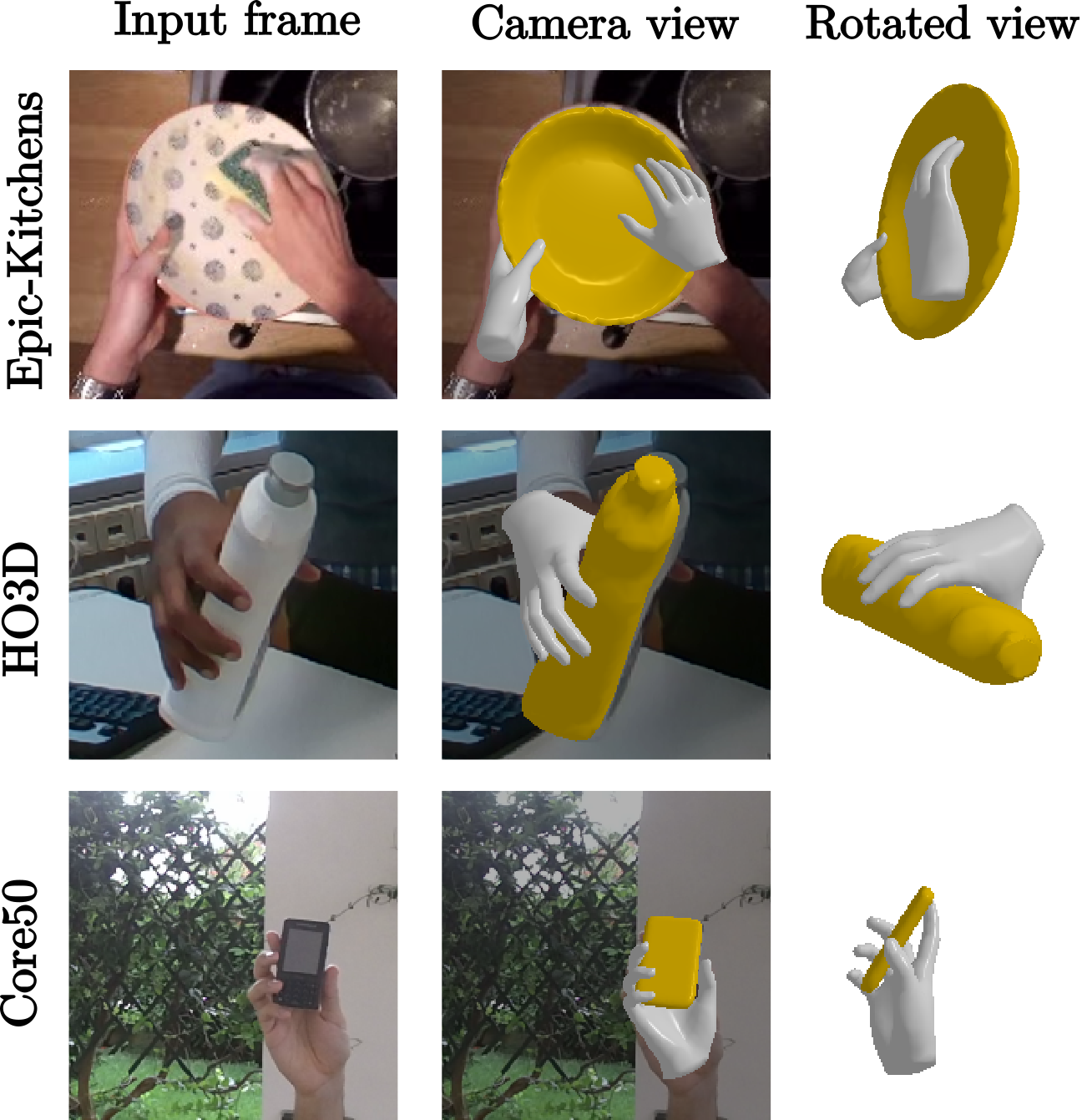}
        \vspace{-.3cm}
    \end{center}
    \caption{\textbf{Hand-object reconstructions:}
        We present an optimization-based method to fit 3D parametric hand model
        and a known object model to 2D estimates. 
        Our method is generic and can be applied across a variety of recent video datasets presenting hand-object interactions~\protect\cite{core50,Damen2018EPICKITCHENS,Hampali_2020_CVPR}.
        Our method can also handle reconstruction of images with left and right hands.
        \vspace{-.4cm}
    }
    \label{fig:teaser}
\end{figure}

Developing an RGB-only method to retrieve 3D hand-object configurations
for diverse objects
would enable scaling up the datasets, %
and help the field move towards in-the-wild scenes.
In this work, we argue for an optimization-based approach for its
robustness across domains. Recent progress in 2D detection of objects
and 3D pose estimation of isolated hands makes it possible to obtain a good initialization
when fitting 3D hand-object poses to these estimates. Nevertheless,
this is still very challenging
due to depth ambiguities, occlusions, noisy 2D estimates and
can result in physically
implausible configurations. In our analysis,
we show improvements over independent fits when considering
a joint hand-object fitting framework with several interaction constraints.
We take inspiration from PHOSA~\cite{phosa}, a recent work showing promising
results for fitting 3D human bodies and objects, relying
on object segmentation and human pose estimation models.

Several approaches propose to explore the complementarity between 
fitting and learning approaches for human body shape estimation.
Fitting for learning has been explored for reconstructing human bodies,
either
semi-automatically~\cite{LassnerUP2017} or
via learning with fitting in-the-loop~\cite{kolotouros2019spin}.
Recently, applying fitting on sign language videos has shown
benefits for training isolated hand reconstruction 
models~\cite{Kulon2020WeaklySupervisedMH}.
\cite{liu2021semi} shows that hand pseudo-labelling and
automatic filtering using spatial and temporal consistency
allows to improve the hand pose estimation branch of a model trained for
joint hand-object pose estimation.
Different from these works, we fit hand \textit{and object} meshes
which represent a more complex scene.
\cite{Hampali_2020_CVPR} presents a similar approach
by fitting hand-object configurations to RGB-D videos to later
use in training;
however, the reliance on a depth sensor significantly
reduces its applicability to unconstrained images.

In this work, we propose a method which reconstructs the hand and manipulated
object from short RGB video clips, assuming an approximate model for the
target object is available.
We investigate the strengths and limitations of our approach in challenging scenarios 
and present results on challenging scenes which are not currently
handled by learning methods for joint hand-object pose estimation.
Our code is publicly available at \url{https://github.com/hassony2/homan}.

Our contributions are the following:
(i)~We propose a fitting-based
approach for hand-object reconstruction
from a video clip;
(ii)~We present a detailed quantitative evaluation
analyzing different components of our optimization method
and compare to learning-based models~\cite{hasson20_handobjectconsist}
on a standard
benchmark~\cite{Hampali_2020_CVPR};
(iii)~We demonstrate qualitatively the capabilities
of our framework to generalize on unconstrained videos and two-hand reconstructions.

\section{Related Work}
\label{sec:related}
We briefly review relevant literature on hand-object reconstruction,
works that employ temporal constraints and hand-object datasets.

\noindent\textbf{Hand-object reconstruction.}
3D pose estimation for
hands~\cite{iqbal2018ECCV,GANeratedHands_CVPR2018,Simon_2017_CVPR,Zimmermann_2017_ICCV,Kulon2020WeaklySupervisedMH} and 
objects~\cite{groueix2018,kato2018renderer,wang2018pixel2mesh,Park_2019_CVPR,Mescheder_2019_CVPR,xiang2018posecnn,labbe2020cosypose,Li2018-fp_deepim}
have often been tackled in isolation~\cite{lepetit2020_handobjadvances}.
Joint reconstruction of hands and objects
has recently received increased 
attention~\cite{hasson19_obman,tekin19_handplusobject,hasson20_handobjectconsist,Hampali_2020_CVPR,Zhe_Radosavovic_2020_arxiv,Doosti_2020_CVPR}.
Hasson et al.~\cite{hasson19_obman} introduces an end-to-end model
to regress MANO~\cite{mano} hand parameters jointly with object mesh vertices deformed from a sphere.
Works of \cite{tekin19_handplusobject,hasson20_handobjectconsist}
assume a known object model and regress a 6DOF object pose instead.
Other methods focus on the grasp synthesis~\cite{GraspingField:3DV:2020,GRAB:2020,Corona_2020_CVPR},
or contact modeling~\cite{contactdb_Brahmbhatt_2019_CVPR,contactpose_Brahmbhatt_2020_ECCV}
given a 3D object.

Interaction constraints are imposed to avoid collisions and
to encourage contact points. To achieve this, competing attraction and repulsion
terms are employed in \cite{hasson19_obman,Corona_2020_CVPR}.
Collision penalization is implemented either with approximate shape 
primitives~\cite{Oikonomidis_2011_ICCV,bogo_2016_ECCV,Kyriazis_2014_CVPR}
or triangle
meshes~\cite{Kyriazis_2013_CVPR,Ballan_2012_ECCV,Tzionas:IJCV:2016,prox,phosa,smplx,hasson19_obman,Hampali_2020_CVPR,Jiang_2020_CVPR}.
Similarly, in this work we impose error terms in
our joint hand-object fitting to favor physically plausible interactions.

Recent work on monocular hand-object reconstruction mostly adopts learning-based CNN models~\cite{hasson19_obman,tekin19_handplusobject,hasson20_handobjectconsist,Hampali_2020_CVPR,KokicKB20,lepetit2020_handobjadvances,liu2021semi}.
Such methods obtain promising results but remain limited to constrained datasets and lack generalization.
Moreover, learning-based methods are typically limited to the predefined number of hands and objects.
In contrast, our method can reconstruct both single and two-hand interactions detected in the image.
Concurrent work of \cite{Zhe_Radosavovic_2020_arxiv} extends
the optimization-based body-object reconstruction method PHOSA~\cite{phosa} 
to perform hand-object fitting. While our method shares
similar optimization components with \cite{Zhe_Radosavovic_2020_arxiv},
it differs by leveraging video data.

\noindent\textbf{Temporal constraints.}
In case of video inputs, temporal constraints have been used for body motion estimation
in the context of neural networks~\cite{Hossain_2018_ECCV,humanMotionKanazawa19},
or optimization~\cite{Arnab_CVPR_2019,2018-TOG-SFV}.
For hands, \cite{Cai_2019_ICCV} proposes a graph convolutional
approach to learn temporal dependencies. Hampali et al.~\cite{Hampali_2020_CVPR}
make use of a temporal consistency term
when fitting hand-object configurations to RGB-D data.
We explore a similar term to obtain temporally smooth fits
to RGB data and initialize the optimization from the previous frame's fit.
\cite{liu2021semi} propose to filter noisy hand reconstructions
by detecting implausible or inconsistent hand poses and jittering predictions
over consecutive frames.

\noindent\textbf{3D label acquisition for hand-object datasets.}
Recent efforts aim to scale up data collection
for \textit{2D} hand-object interactions~\cite{Shan20}.
Due to the difficulty of annotating 3D in images,
several approaches for \textit{automatic} label acquisition have been proposed.
\cite{hasson19_obman} introduces ObMan, a synthetic dataset obtained by rendering MANO
hand model~\cite{mano} automatically grasping ShapeNet objects~\cite{ShapeNet}.
The dataset of~\cite{fphab_hernando_2018_cvpr}
utilizes visible magnetic sensors while~\cite{Pham_2018_TPAMI} rely on force sensors which limit the range of posible grasps.
\cite{Hampali_2020_CVPR}~presents the HO-3D dataset where labels result
from automatic fits to RGB-D data.
\cite{contactpose_Brahmbhatt_2020_ECCV} collects a dataset for 3D printed object models
in a restricted capture setup.
Very recently, \cite{chao:cvpr2021} and \cite{kwon2021h2o} propose to collect annotations using hybrid methods to collect reliable labels.
In this work, we use the HO-3D benchmark for quantitative evaluations.
We also show qualitative results for videos in-the-wild where 3D ground truth is not available.

\section{Automatic labelling of 3D hands and objects}
\label{sec:method}

\begin{figure*}
\begin{center}
\includegraphics[width=1\textwidth,bb=0 0 16cm 3cm]{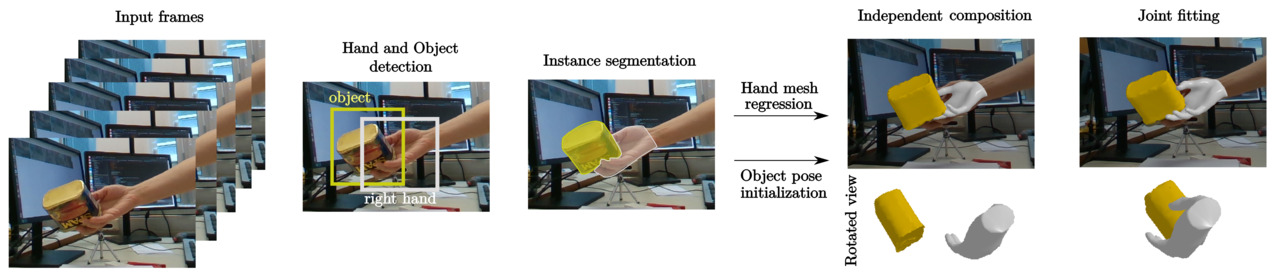}
\vspace{-.2cm}
\end{center}
   \caption{\textbf{Joint hand-object fitting:}
   We independently initialize the hand and object poses
   based on 2D detections and segmentations. We refine this
   configuration with interaction-based constraints to obtain
   our final joint fitting.
   }
\label{fig:pipeline}
\end{figure*}

We first describe the optimization-based fitting procedure,
consisting of estimating 2D detections  
(Section~\ref{subsec:evidence}),
initializing 3D hand and object poses 
(Section~\ref{subsec:independent}),
and joint fitting (Section~\ref{subsec:fitting}).
An overview can be seen in Figure~\ref{fig:pipeline}.

Our method takes a video of hand-object manipulation as input.
We assume that an exact or approximate object model representing the manipulated object is provided and use the ground-truth camera intrinsic parameters when available.

\subsection{Obtaining 2D hand-object evidence}
\label{subsec:evidence}

\noindent\textbf{2D hand and object detection.}
For each video, 
the first step is to detect initial 2D bounding boxes.
We use a recent hand and manipulated object detector~\cite{Shan20} to extract object and hand bounding boxes in each frame. The hands are predicted with left or right side labels.
When the predicted boxes do not match the known properties of 
the dataset in terms of object presence, hand number or hand sides, we discard the detections for the given frame, and recover detections through tracking in a subsequent step.

\noindent\textbf{Tracking.}
We apply an off-the-shelf 2D bounding box tracker~\cite{motpy} which relies on Kalman filtering~\cite{labbe2014} to extract hand and object tracks from the noisy per-frame detections.
This step allows to recover missed or discarded detections, and produces hand and object bounding box candidates for the full video clip.

For the Epic-Kitchens dataset, the real number of visible hands is unknown.
We automatically select video clips for which at least one object and one hand track extend over more then 20 consecutive frames after tracking.

\noindent\textbf{Segmentation.} 
The key image evidence we rely on for fitting
is 2D segmentation.
We extract instance masks $\hat{\mathcal{M}}_{obj}$ for each tracked object detection using the instance segmentation head of the PointRend~\cite{kirillov2019pointrend}.
Similar to PHOSA~\cite{phosa}, we use a model pretrained on the COCO~\cite{coco} dataset.
However, while \cite{phosa} fits %
objects among the COCO categories, our target everyday objects are often not
present among the COCO classes. 
For each object detection, we use the
mask associated to the
highest class activation of the PointRend instance classifier.
We observe this class-agnostic approach
to perform well in most cases.
To account for hand occlusions, we extract the COCO masks associated to the \textit{person} class $\hat{\mathcal{M}}_{hand}$ for the tracked hand boxes, see the \if\sepappendix1{Section~B}
\else{Section~\ref{app:sec:super2d}}
\fi of the supplemental material 
for additional details.

\subsection{Independent pose initialization}
\label{subsec:independent}

\noindent\textbf{Hand initialization.}
We employ the recent publicly available hand pose estimator FrankMocap~\cite{rong2020frankmocap} to estimate the initial hand articulated poses, as well as the hand location and scale in pixel space.
We recover an estimated depth using the world scale of the hand and the exact intrinsic camera parameters when available.
When the exact camera intrinsic parameters are unknown, we approximate the focal length given the specifications of the camera, and assume the central point is at the center of the pixel image.

\noindent\textbf{Object initialization.}
We use the 2D object segmentation to initialize the object pose for the 3D model associated to the target video clip.
To obtain pose candidates for the first video frame, we sample random rotations uniformly in $\mathcal{SO}(3)$ and use the radius of the instance bounding box to estimate the $3D$ center of the provided mesh in the first frame.
We update the object depth 
minimizing the difference between the \textit{diagonals} of the projected model's tight bounding box and the predicted detection in the least squared sense. 
We then compute a translation update in the camera plane, minimizing the difference between the \textit{centers} of the projected object model and the detection given a fixed depth.
We repeat these two steps until convergence to recover an estimate of the object translation.
We further optimize the object pose using differentiable rendering.
We optimize the object pose with a pixel squared loss on the difference
between the differentiably rendered object mask and the PointRend~\cite{kirillov2019pointrend}
object segmentation following PHOSA~\cite{phosa}.
As for PHOSA, our object loss is hand-occlusion aware: only pixels which are not associated
with the person label are taken into account. 
For each subsequent frame, we use the object pose from the previous frame as initialization and
further refine it with differentiable rendering as described above.
This process results in as many candidate motion initializations as there are candidate object poses. In practice, the number of candidate initializations is empirically set to 50.

We select the object motion candidate for which the average $IoU$ score between the rendered mask and the target occlusion mask %
is highest.

\subsection{Joint fitting}
\label{subsec:fitting}

Independent hand-object fits are often inaccurate
and do not take into account interaction-based constraints.
We %
refine the initial hand-object poses %
leveraging both coarse and fine-grained manipulation priors.

\vspace{0.2cm}
\noindent\textbf{Optimized parameters.}
The goal of our fitting is to find the optimal hand and object pose parameters for a sequence of $T$ consecutive frames. For each frame, we optimize 3D translations $D_{hand}$, $D_{obj}$, 3D global rotations $R_{hand}$, $R_{obj}$ as well as $\theta$ hand pose parameters. Additionally, we optimize a shared hand scale $s_{hand}$.

We optimize the articulated MANO~\cite{mano} model in the latent pose space $\theta$. Given the $\theta$ pose parameters, the MANO model differentiably outputs 3D hand vertex coordinates centered on the middle metacarpophalangeal joint $\mathcal{V}_{hand}^{c} = \textrm{MANO}(\theta)$.
Following~\cite{GRAB:2020}, we use a pose latent subspace of size $16$.

We optimize the hands and object rotation $R_{hand}, R_{obj}$ using the 6D continuous rotation representation~\cite{Zhou_2019_CVPR} and optimize the 3D translation $D_{hand}, D_{obj}$ in metric camera space.
When we use approximate object meshes, we additionally optimize a scalar scaling parameter $s_{obj}$ which allows the object's size to vary.
We also allow hand vertices to scale
by a factor $s_{hand}$ which is shared across the $T$ frames.
The hand vertices in camera coordinates $\mathcal{V}_{hand}^{3d}$ are estimated as following:
\begin{equation}
    \mathcal{V}_{hand}^{3d} = s_{hand} (R_{hand}  \mathcal{V}_{hand}^{c}) + D_{hand}.
\end{equation}

The object vertices $\mathcal{V}_{obj}^{3d}$  are estimated as a rigid transformation of canonically oriented model vertices $\mathcal{V}_{obj}^{c}$:
\begin{equation}
    \mathcal{V}_{obj}^{3d} = s_{obj} (R_{obj} \mathcal{V}_{obj}^{c}) + D_{obj}.
\end{equation}
Next, we describe the individual error terms that we
minimize during fitting.

\noindent
\textbf{Object silhouette matching ($\mathcal{L}_{obj}$).} We use a differentiable renderer~\cite{kato2018renderer} to render the object mask $\mathcal{M}_{obj}$ and compare it to the reference segmentation mask $\hat{\mathcal{M}}_{obj}$. This error term is occlusion-aware as in~\cite{phosa}. No penalization occurs for the object silhouette being rendered in pixel regions $\hat{\mathcal{M}}_{hand}$ where hand occlusions occur. We write this error as:
\begin{equation}
    \mathcal{L}_{obj} = ||(1 - \hat{\mathcal{M}}_{hand}) \circ (\mathcal{M}_{obj} - \hat{\mathcal{M}}_{obj}) ||^2_2\textbf{}
\end{equation}

\vspace{0.1cm}
\noindent
\textbf{Projected hand vertices %
($\mathcal{L}^{v2d}$).}
We constrain the hand position by penalizing projected vertex offsets from the initial vertex pixel locations $\mathcal{V}^{hand}_{2d}$ predicted by FrankMocap~\cite{rong2020frankmocap}. To compute the current 2D vertex locations, we project the MANO~\cite{mano} vertices $\mathcal{V}^{3D}_{hand}$ to the pixel plane using the camera projection operation $\Pi$. This error is written as:
\begin{equation}
    \mathcal{L}_{v2d} = || \Pi(\mathcal{V}^{3D}_{hand}) - \hat{\mathcal{V}}_{hand}^{2d} ||^2_2
\end{equation}

\vspace{0.1cm}
\noindent
\textbf{Hand regularization ($\mathcal{L}_{pca}$).} Given that we optimize the \textit{articulated} hand pose, we regularize the optimized hand pose.
As in~\cite{rong2020frankmocap,hasson19_obman,Boukhayma_2019_CVPR}, we apply a mean square error term to the PCA hand components $\mathcal{L}_{pca} = || \theta ||^2_2$ to bias the estimated hand poses towards statistically plausible configurations.

\noindent
\textbf{Scale ($\mathcal{L}_{scale}$).} Similarly to PHOSA\cite{phosa}, when we allow the elements in the scene to scale, we penalize deviations from category-level average dimensions.

\vspace{0.1cm}
\noindent
\textbf{Smoothness ($\mathcal{L}_{smooth}$).} We further leverage a simple smoothness prior over the $T$ sampled frames $\mathcal{L}_{smooth} = \sum_{t=1}^{T-1} ||\mathcal{V}^{3D}_{t+1} - \mathcal{V}^{3D}_{t}||^2_2 $ which encourages minimal 3D vertex variances across neighboring frames for both hands and objects as in~\cite{Hampali_2020_CVPR}.

\vspace{0.1cm}
\noindent
\textbf{Coarse interaction ($\mathcal{L}_{centroid}$).} Following~\cite{phosa}, we penalize the squared distance between hand and object centroids when the predicted hand and object boxes overlap to encode a coarse interaction prior.
As we assume the object scale to be provided, this error only impacts the rigid hand pose, effectively attracting the hand towards the interacted object. 
In case of multiple hands, all overlapping hand-object pairs of meshes are considered.

\vspace{0.1cm}
\noindent
\textbf{Collision ($\mathcal{L}_{col}$).} We rely on a recent collision penalization term introduced to enforce non-interpenetration constraints between multiple persons in the context of body mesh estimation~\cite{Jiang_2020_CVPR}.
The collision error $\mathcal{L}_{col}$ is computed for each pair ${k, l}$ of estimated meshes. We compute $L_{col}^{k, l} = \sum_{i} {\Phi_{k}(\mathcal{V}_{l}^i)}$. Where $\Phi_{k}$ is the negative truncated signed distance function (SDF) associated to the mesh ${k}$, $\Phi_{k}(\mathcal{V}) = max(0, - SDF(\mathcal{V}))$.
\begin{equation}
    \mathcal{L}_{col} = \sum_{{k, l}} \mathcal{L}_{col}^{k, l} 
\end{equation}
This formulation allows to handle any number of visible hands and objects in the scene.

\vspace{0.1cm}
\noindent
\textbf{Local contacts.} Hands interact with objects by establishing surface contacts without interpenetration. We experiment with the hand-object heuristic introduced by~\cite{hasson19_obman}. We re-purpose this loss which has been introduced in a learning framework to our optimization setup. This additional term encourages the contacts to occur at the surface of the object by penalizing the distance between hand vertices the closest object vertex for hand vertices in the object's vicinity. We refer to the supplemental material
\if\sepappendix1{Section~C}
\else{Section~\ref{app:sec:implem-details}}
\fi
for additional details.

The final objective $\mathcal{L}$ is composed of a weighted sum of the previously described terms, where the weights are empirically set to balance the contributions of each error term.

\begin{equation}
\label{eq:objective}
\begin{gathered}
     \mathcal{L} = \lambda_{obj}  \mathcal{L}_{obj} + \lambda_{v2d} \mathcal{L}_{v2d}    \\
     + \lambda_{pca} \mathcal{L}_{pca} + \lambda_{scale} \mathcal{L}_{scale} + \lambda_{smooth} \mathcal{L}_{smooth} \\ 
     + \lambda_{centroid} \mathcal{L}_{centroid} + \lambda_{local} \mathcal{L}_{local}  + \lambda_{col} \mathcal{L}_{col}
\end{gathered}
\end{equation}

While~\cite{phosa} adapt the weights for their optimization for each object categories, we fix the weight parameters  empirically and keep them constant across all experiments. We refer to 
\if\sepappendix1{Section~C}
\else{Section~\ref{app:sec:implem-details}}
\fi of the supplemental material 
for exact weight values and additional implementation details.

\section{Experiments}\label{sec:experiments}

We first define the evaluation metrics (Section~\ref{subsec:metrics}) and the datasets 
(Section~\ref{subsec:datasets})
used in our experiments.
Then, we provide an ablation to measure the contribution
of each of our optimization objective terms (Section~\ref{subsec:errorterms}).
We investigate the sensitivity of our approach to
the quality of the 2D estimates (Section~\ref{subsec:sensitivity}).
Next, we compare our approach to the state of the art (Section~\ref{subsec:sota}).
Finally, we provide qualitative results for in-the-wild examples (Section~\ref{subsec:inthewild}).

\subsection{Metrics}\label{subsec:metrics}

The structured output for hand-object reconstruction is difficult to evaluate with a single metric.
We therefore rely on multiple evaluation measures.

\noindent\textbf{Object metrics.} We evaluate our object pose estimates by computing the average vertex distance. 
Common objects such as bottles and plates often present plane and revolution symmetries.
To account for point matching ambiguities, we further report the standard pose estimation 
average closest point distance (add-s)~\cite{xiang2018posecnn}.

\noindent\textbf{Hand metrics.}
We follow the standard hand pose estimation protocols~\cite{Hampali_2020_CVPR,Zimmermann_2019_ICCV} and report 
the procrustes-aligned hand vertex error and F-scores. 
We compare the hand joint predictions using average distances after scale and translation alignment.
When investigating the results of the joint fitting, we additionally report the average hand vertex distances without alignment.

\noindent\textbf{Interaction metrics.}
\noindent
\textit{Penetration depth (mm)}: We report the maximum penetration depth between the hand and the object following previous work on hand-object interactions~\cite{GraspingField:3DV:2020,hasson19_obman,hasson20_handobjectconsist,contactpose_Brahmbhatt_2020_ECCV}.
\noindent
\textit{Contact (\%):} We also report the contact percentage following~\cite{GraspingField:3DV:2020}.
When ground truth contact binary labels are available, we report contact accuracy, as we further detail in the supplemental material
\if\sepappendix1{Section~C}
\else{Section~\ref{app:sec:implem-details}}
\fi.

\subsection{Datasets}\label{subsec:datasets}

\noindent\textbf{HO-3D}~\cite{Hampali_2020_CVPR} is the reference dataset which provides accurate hand-object annotations during interaction for marker-less RGB images. The users manipulate $10$ objects from the YCB~\cite{ycb} dataset, for which the CAD models are provided. The ground-truth annotations are obtained by fitting the hand and object models to RGB-D evidence which assumes limited hand motion. We present results on the standard test set composed of $13$ videos with a total of $11525$ frames depicting single-hand object manipulations. 

\noindent\textbf{Core50}~\cite{core50} contains short sequences of unannotated images of hands manipulating $50$ object instances from $10$ everyday object categories such as cups, light bulbs and phones.
We manually associate $26$ objects with approximately matching 3D object models from the ShapeNet dataset~\cite{ShapeNet}.
We further annotate hands being left or right for each of the $11$ video sequences available for each object, resulting in $286$ video clips and $86k$ frames.

\noindent\textbf{Epic Kitchens}~\cite{Damen2018EPICKITCHENS} is an unscripted dataset which has been collected without imposing constraints or equipment beyond a head-mounted camera. In contrast to existing datasets for which 3D information is available, it therefore presents natural hand-object interactions.
This dataset is however densely annotated with action labels which include the category of the object of interest.
We focus on a subset of common object categories: cups, plates, cans, phones and bottles which are involved in a total of $3456$ action video clips.
For each object category, we associate an object model from the CAD ShapeNet~\cite{ShapeNet} database.
We assume that the target manipulated object is the one with the longest track in the associated action clip.

\subsection{Contribution of error terms in fitting}
\label{subsec:errorterms}

\begin{figure}[t]
\begin{center}
   \includegraphics[width=0.99\linewidth,bb=0 0 23cm 23cm]{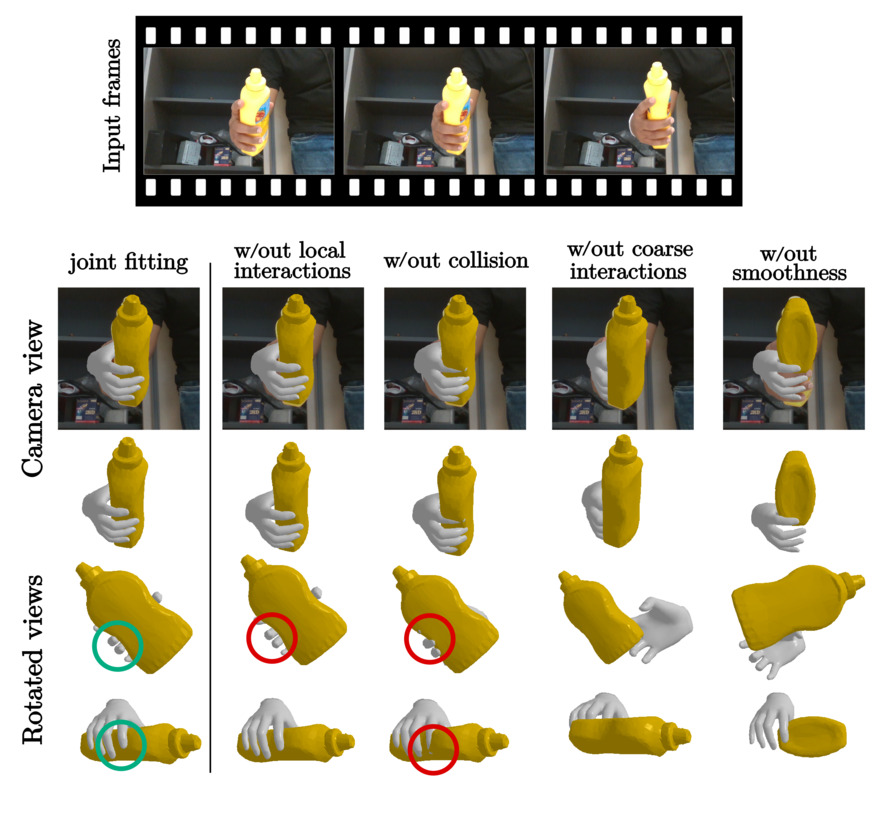}
   \vspace{-.5cm}
\end{center}
   \caption{\textbf{Effect of error terms:} Qualitative analysis showing the effects of the various
   error terms for the hand-object reconstruction accuracy on the HO-3D dataset. We highlight visual evidence of local corrections attributed to the local interaction~\protect\cite{hasson19_obman} and collision~\protect\cite{Jiang_2020_CVPR} terms.}
\label{fig:phosaqualitative}
\end{figure}

As explained in Section~\ref{sec:method}, our method introduces
several error terms which determine the final reconstruction quality.
We evaluate the contribution of the main error terms on the HO-3D 
benchmark~\cite{Hampali_2020_CVPR} in Table~\ref{tab:phoman_ablation}.
We validate that our joint reconstruction outperforms the naive composition baseline, which is 
obtained by separately fitting the object to the occlusion-aware object mask and the hand using 
the hand-specific terms $\mathcal{L}_{v2d}$ and $\mathcal{L}_{pca}$.
When fitted independently, the scale-depth ambiguity prevents an accurate estimate of the hand distance.
As we use the ground-truth object model, the 3D object pose can be estimated without ambiguity 
using the camera intrinsic parameters.
Joint fitting improves the 3D pose estimates using both smoothness and interaction priors.
We observe that the coarse interaction prior is critical towards improving the absolute hand 
pose.
When removing this error term, the hand pose error increases two-fold from $8.9$ to $17.1\textrm{cm}$.
We observe that the temporal smoothness term,
while simple, provides a strong improvement to both the hand and object pose estimates.
Leveraging information across neighboring video frames reduces the errors by $25\%$ and $38\%$ 
for the hand and object, respectively. 
While the local interaction and collision penalization terms only marginally change the hand and object reconstruction scores, their impact can be quantitatively observed in the interaction 
metrics.
The collision penalization terms reduce the average penetration depth by a large factor
($10.2\textrm{mm}$ vs $2.4\textrm{mm}$).
The local interaction term~\cite{hasson19_obman} reduces both the interpenetration depth and the contact percentage, which is defined as either exact surface contact or interpenetration between the hand and the object. Qualitatively, we observe that this term produces local corrections in the vicinity of estimated contact points.
Including all error terms results in more plausible grasps, which
we illustrate in Figure~\ref{fig:phosaqualitative} qualitatively.

\begin{table*}
    \centering
    \resizebox{0.99\linewidth}{!}{
        \begin{tabular}{l|cc|cc|cc}
        \toprule
        & \multicolumn{2}{|c|}{Hand} & \multicolumn{2}{|c|}{Object} & \multicolumn{2}{|c}{Interaction} \\\cline{2-7}
              & vertex mean & mepe   & vertex mean & add-s & pen. depth & contact \\ 
              &   distance (cm)$\downarrow$  & aligned (cm) $\downarrow$ & distance (cm) $\downarrow$ &  (cm) $\downarrow$ &  (mm) $\downarrow$  & \% \\
        \hline

        indep. composition & 26.2  & 5.2 &  12.1 & 7.7 & 3.2 & 25.8 \\ 
        \rowcolor{aliceblue}
        joint fitting  &  8.6 & 5.4 & 8.0 & 3.8 & 2.8 & 77.5 \\
        \quad w/out local interactions & 8.5 & 5.4 & 8.0 & 3.8  & 2.4 & 72.3 \\
        \quad w/out collision & 8.9 & 5.4 & 8.0 & 3.8 & 10.2 & 80.5 \\
        \quad w/out coarse interaction & 17.1 & 5.3  & 8.1 & 3.8 & 1.9  & 59.4 \\ 
        \quad w/out smoothness & 11.4 & 5.6 & 12.8 & 8.3 & 3.0 & 79.1  \\   
        \bottomrule
        \end{tabular}
    }
    \caption{\textbf{Contribution of error terms:} We show benefits of the \textit{joint} modeling for hand-object interactions by the increased reconstruction accuracy when compared to independent hand and object composition on the HO-3D~ \protect\cite{Hampali_2020_CVPR} dataset. Our smoothness and interaction terms impose additional constraints which improve the final hand-object pose reconstructions.}
    \label{tab:phoman_ablation}
\end{table*}

While ground truth 3D poses are hard to annotate for generic videos with hand-object manipulations, interaction metrics such as penetration depth can be directly computed from the predicted reconstructions.
As the Core50~\cite{core50} dataset presents only videos in which the object is actively manipulated by the hand, we can additionally report contact accuracy as proxies  to evaluate the quality of the reconstructed grasps.
We report these two metrics on the Core50 dataset in Table~\ref{tab:core50_contact_accuracy} and confirm the benefit from our joint fitting approach.
Using joint fitting significantly increases the contact accuracy from 7.3\% to 89.5\%, while only increasing the average penetration depth by $0.6\textrm{mm}$. 

\begin{table}
\begin{tabular}{l|cc|cc}
\toprule
Dataset    & \multicolumn{2}{c|}{Contact Accuracy (\%)~$\uparrow$} & \multicolumn{2}{c}{Pen. Depth (mm) $\downarrow$} \\
      & independent & joint & independent & joint \\ 
\midrule
Core50  &  7.3 & \textbf{89.5} & \textbf{0.6} & 1.2 \\
\bottomrule
\end{tabular}
\caption{\textbf{Results on Core50:} Interaction errors for hand-object fits obtained on the Core50 dataset. We observe significantly improved contact accuracy with
joint fitting over independent fits at the expense of a minor cost of a $0.6\textrm{mm}$  increase in penetration.}
\label{tab:core50_contact_accuracy}
\end{table}

\subsection{Sensitivity to estimated 2D evidence}
\label{subsec:sensitivity}

\begin{figure}[t]
\begin{center}
   \includegraphics[width=0.99\linewidth,bb=0 0 25cm 8cm]{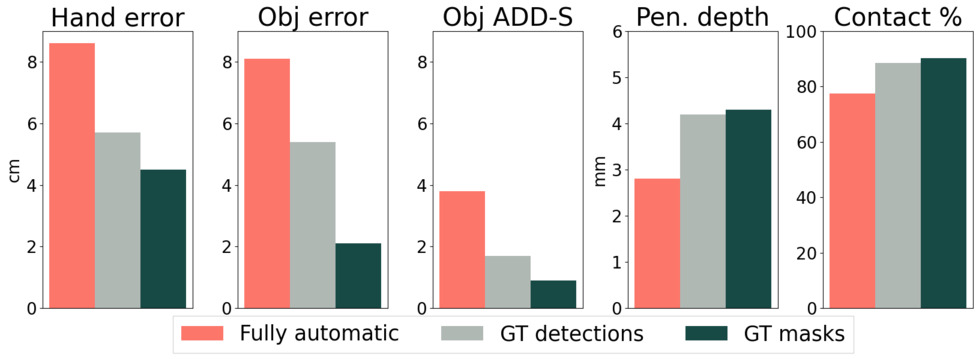}
   \vspace{-.3cm}
\end{center}
   \caption{\textbf{Sensitivity to 2D detections:} Dependence of our 3D reconstruction on the accuracy of the 2D evidence by running our method with ground truth (GT) hand and object detections and ground truth object masks for the HO-3D dataset~\protect\cite{Hampali_2020_CVPR}.}
\label{fig:abl2d}
\end{figure}

Our method makes use of generic models for detection and mask estimation and would directly benefit from more accurate detection, 2D hand pose estimation and instance segmentation models.
The reliance of our method on 2D cues therefore allows it to benefit from additional efforts in 2D image annotation which is simpler compared to 3D annotation in practice.
We investigate the dependence of our method on the quality of the available 2D evidence.
To investigate the expected improvements our method could gain from stronger object detections, instead of using noisy detections, we use the hand and object ground truth bounding boxes provided for the HO-3D test set.
We observe in Figure~\ref{fig:abl2d} the improvements we obtain from using the ground-truth detections.
Both hands and objects benefit from more accurate detections, improving by $2\textrm{cm}$ when compared to the tracking-by-detection estimates. 
To investigate the errors which come from using noisy approximate instance masks, we render the object and hand ground truth masks and use them to guide our optimization.
By relying on 2D information, our approach  suffers from limitations such as depth ambiguities which can result from fitting to image segmentation masks.
Object asymmetries which rely on color information can also be hard to resolved during fitting.
We observe that using ground truth hand and object masks allows to further decrease the 3D pose errors.
We note that the object error decreases to $2\textrm{cm}$ and below $1\textrm{cm}$ when comparing distances to closest points. When the object model is available, our joint fitting method produces highly accurate object poses in the presence of accurate 2D evidence.

\subsection{State-of-the-art comparison}
\label{subsec:sota}

Recent efforts for joint hand-object pose estimation in camera 
space~\cite{hasson19_obman,tekin19_handplusobject,hasson20_handobjectconsist} have focused on 
direct bottom-up regression of 3D poses.
We compare the performance of our fitting approach to recent learning-based methods for joint 
hand-object reconstruction~\cite{hasson20_handobjectconsist,liu2021semi} in Table~\ref{tab:ho3d-sota}.

\begin{table}
\resizebox{0.99\linewidth}{!}{
\begin{tabular}{l|cccc}
\toprule
        & mesh & F-score & F-score & generalizes to \\ 
Method    &  error $\downarrow$ & @5mm $\uparrow$ & F@15mm $\uparrow$  & unseen objects \\ 
\midrule
Liu 2021~\cite{liu2021semi} & \textbf{0.95} & \textbf{0.96} & \textbf{0.53} & \xmark \\
Hampali 2020~\cite{Hampali_2020_CVPR}    & 1.06 & 0.51 & 0.94 & \xmark \\
Hasson 2020 ~\cite{hasson20_handobjectconsist}  & 1.14 &  0.42 & 0.93 & \xmark  \\
\midrule
Joint fitting & 1.47 & 0.39 & 0.88 & \cmark \\
\bottomrule
\end{tabular}
}
\caption{\textbf{State-of-the-art-comparison:}
We compare the hand performance of the single-view pose estimation baselines.
Note that the reported results for~\cite{Hampali_2020_CVPR} output hand meshes only.
Hasson 2020~\protect\cite{hasson20_handobjectconsist}, Liu 2021~\protect\cite{liu2021semi} and our method predict the hand-object meshes jointly. All methods are trained only on the real images from the HO-3D training split and evaluated on the official test split through an online submission\protect\footnotemark. The hand mesh error is reported after procrustes alignment.} 
\label{tab:ho3d-sota}
\end{table}
\footnotetext{\href{https://competitions.codalab.org/competitions/22485}{https://competitions.codalab.org/competitions/22485}}

While these learnt methods produce more accurate hand predictions, their object predictions are 
\textit{instance specific}.
As a direct consequence, the methods do not generalize to new objects at test time.
In contrast, our generic fitting method performs equally well across the seen and unseen objects of the HO-3D test split, see Table~\ref{tab:unseen_objects}.

\begin{figure*}[t]
\begin{center}
   \includegraphics[width=0.99\linewidth,bb=0 0 47cm 30cm]{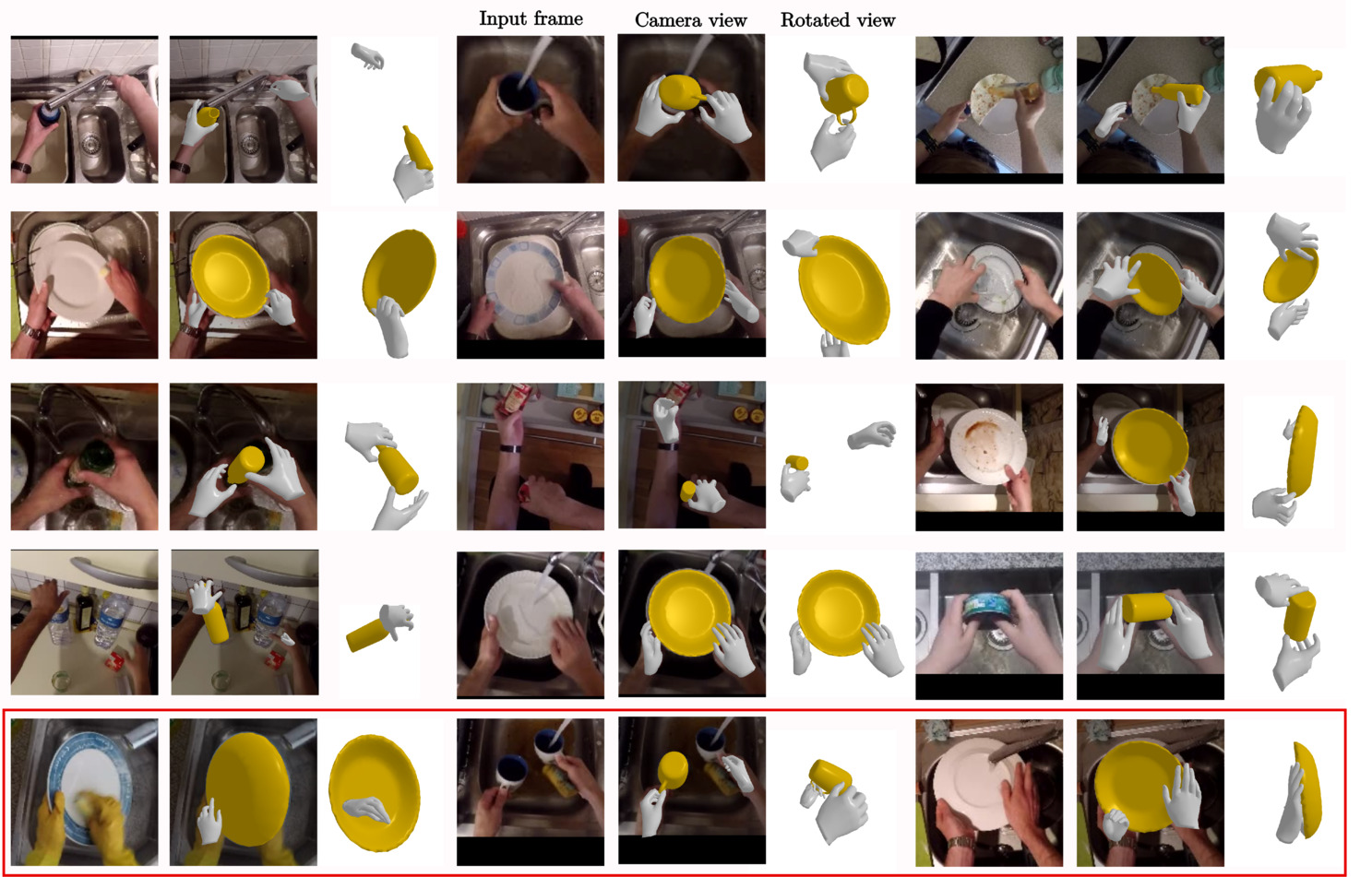}
\end{center}
    \vspace{-.5cm}
   \caption{\textbf{In-the-wild reconstructions:} Our results on natural hand-object manipulations of the Epic-Kitchens dataset~\protect\cite{Damen2018EPICKITCHENS}. We present several success and failures of our method on the challenging Epic-Kitchens dataset. We highlight typical failure modes for our method, in particular, object orientation errors resulting from depth ambiguity.
   We observe that our fitting method recovers plausible interactions across different object categories and hand-object configurations. }
\label{fig:epicmanipulation}
\end{figure*}

\begin{table}
\resizebox{0.99\linewidth}{!}{
\begin{tabular}{l|cc|cc|cc}
\toprule
                &  \multicolumn{4}{c|}{Object} & \multicolumn{2}{c}{Hand} \\
                & \multicolumn{2}{c|}{vertex dist (cm)  $\downarrow$}  & \multicolumn{2}{c|}{add-s (cm)  $\downarrow$} & \multirow{2}{*}{mepe $\downarrow$}  &  aligned \\ 
                         \cline{2-5}
                 &Seen & Unseen  &  Seen & Unseen &  & mepe $\downarrow$ \\
\hline
Ours & 8.0 & 8.1 & 4.0 & 3.3  & 8.6 & 5.4  \\
~\cite{hasson20_handobjectconsist} & 6.7 & 10.7 & 2.2 & 3.6  & \textbf{5.5} & \textbf{3.7} \\ 
\bottomrule
\end{tabular}
}
\caption{\textbf{Unseen objects:} Vertex errors (cm) for estimated hand and object meshes. Compared to~\protect\cite{hasson20_handobjectconsist}, our method performs similarly across seen and \textit{unseen} objects.}
\label{tab:unseen_objects}
\end{table}

\begin{figure}%
\begin{center}
   \includegraphics[width=0.99\linewidth,bb=0 0 32cm 15cm]{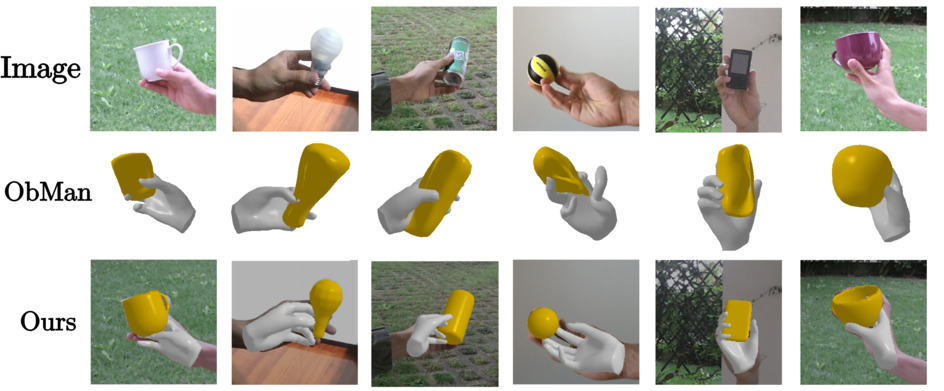}
\end{center}
    \vspace{0cm}
   \caption{\textbf{Comparison with \protect\cite{hasson19_obman}:} Qualitative comparison of our
   fits to the ObMan-trained model~\protect\cite{hasson19_obman} estimations on the Core50 dataset.%
   While our model requires an approximate mesh to be provided, it generalizes to objects of arbitrary topology.}
\label{fig:core50qualitative}
\end{figure}

\subsection{In-the-wild 3D hand-object pose estimation}
\label{subsec:inthewild}

We test the limits of our approach and
showcase the strength of our method by comparing qualitatively to a model trained for joint hand-object reconstruction~\cite{hasson19_obman}.
Their model estimates the shape of the object by deforming a sphere and therefore does not depend on known object models.
However, given this object topology restriction, this method has limited expressivity.
It can only capture a subset of all object shapes which excludes everyday objects such as mugs or cups, see Fig.~\ref{fig:core50qualitative}.
In comparison, while our method makes stronger assumptions by relying on an approximate object model, it is applicable to any everyday objects for which an approximate mesh can be retrieved without further limitations.
Additionally, while the manipulation reconstruction from~\cite{hasson19_obman} estimates the grasp relative to the root joint of the hand, our method outputs image-aligned predictions.

We further show that our method can be applied to the challenging Epic-Kitchens dataset~\cite{Damen2018EPICKITCHENS} which presents natural manipulation of common objects, see Fig.~\ref{fig:epicmanipulation}.
Note that our objective~(\ref{eq:objective}) is not restricted to a single hand-object pair and naturally generalizes to multiple hands and objects.
To handle scenes with two hands in~\cite{Damen2018EPICKITCHENS} we optimize  (\ref{eq:objective}) with pairwise losses defined for both detected hands and the detected object.
We show results of two-hand manipulations which represent the majority of examples in the target dataset.
While we observe cases of depth ambiguity, especially with almost plannar objects such as plates, we show that our method can recover plausible reconstructions for several object categories across a variety of hand poses.

\section{Conclusions}
\label{conclusions}

We present a robust approach
for fitting 3D hand-object
configurations to monocular
RGB videos.
Our method builds on
estimates from neural network models for detection, object
segmentation and 3D hand pose estimation
trained with full supervision.
Due to the lack of supervision
at similar scale for 3D hand-object interactions,
we opt for a fitting-based
approach and demonstrate 
advantages on several datasets.
A key limitation of current methods
that estimate 6DOF object pose is the reliance
on known object models. Future work
will consider automatic object recovery
to fully automate the hand-object reconstruction process.

\paragraph{Acknowledgments.} This work was funded in part by the MSR-Inria joint lab, the French government
under management of Agence Nationale de la Recherche as
part of the ``Investissements d’avenir'' program, reference
ANR19-P3IA-0001 (PRAIRIE 3IA Institute), by Louis
Vuitton ENS Chair on Artificial Intelligence, as well as gifts from Google and Adobe.
This work was granted access to the HPC resources of IDRIS under the allocation 2020-AD011012004R1 made by GENCI.

{\small
\bibliographystyle{ieee_fullname}
\bibliography{ms}
}

\renewcommand{\thefigure}{A.\arabic{figure}} %
\setcounter{figure}{0} 
\renewcommand{\thetable}{A.\arabic{table}}
\setcounter{table}{0} 

\appendix
\section*{APPENDIX}

We first present additional qualitative results for the HO-3D~\cite{Hampali_2020_CVPR} dataset (Section~\ref{app:sec:ho3d-quali}).
Next, we analyze the 2D detection and segmentation performance of the general-purpose estimators on the HO-3D~\cite{Hampali_2020_CVPR} dataset (Section~\ref{app:sec:super2d}).
Finally, we provide additional implementation details for our joint hand-object optimization (Section~\ref{app:sec:implem-details}).

\section{Qualitative results for HO-3D dataset}
\label{app:sec:ho3d-quali}

\begin{figure*}[t]
  \includegraphics[width=0.99\linewidth,bb=0 0 12cm 3cm]{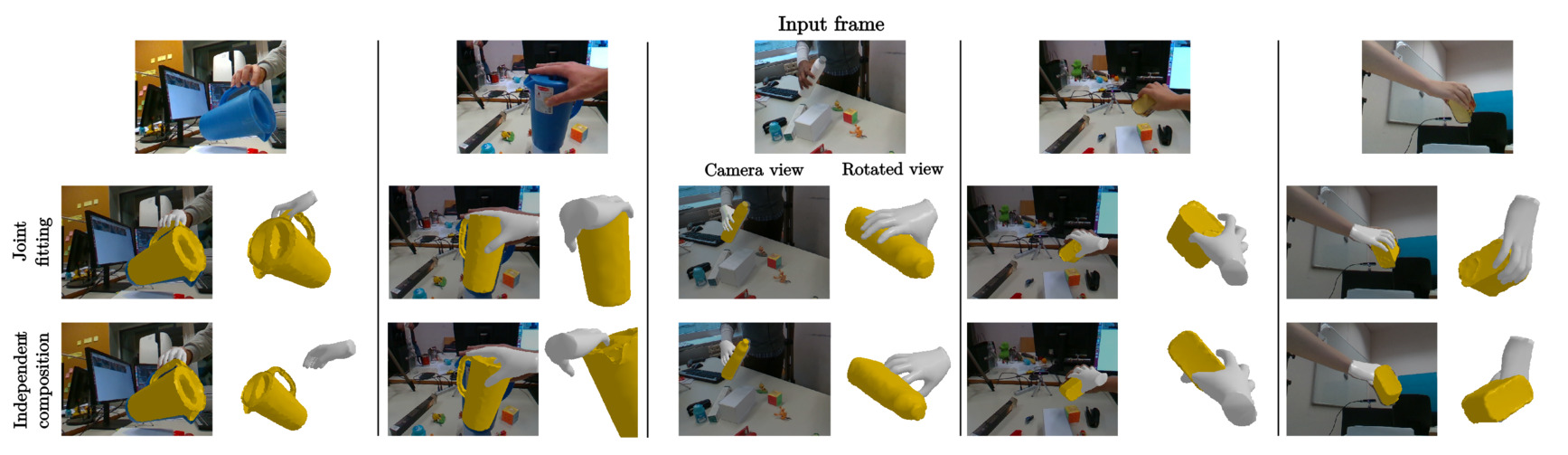}
  \caption{Reconstructions obtained from independent composition and joint fitting on HO-3D~\cite{Hampali_2020_CVPR} dataset. We observe that enforcing hand-object interaction constraints allows our method to recover plausible grasps and significantly improves upon the naive independent composition baseline.}
  \label{fig:qualitative_ho3d}
\end{figure*}

In Figure~\ref{fig:qualitative_ho3d}, we compare qualitatively on the HO-3D test set
our joint fitting method with independent composition, which results from fitting the object and the hand without interaction constraints independently on each frame.
We observe that the joint fitting allows to recover improved configurations across various test objects and consistently improves upon independent composition. 

\section{2D evidence accuracy.}
\label{app:sec:super2d}

\subsection{Evaluation of hand and object detection}
\label{subsec:detect2d}

We assess the performance of our procedure to recover the detections and segmentation masks for the hands and manipulated objects.

We use the 100 Days of Hands (100DOH) hand-object detector~\cite{Shan20} to predict hand and object bounding boxes.
Following~\cite{Shan20}, we evaluate hand detection accuracy by reporting average precision (AP) %
for predicted hand and object boxes and set the intersection over union (IoU) threshold to 0.5 on the HO-3D~\cite{Hampali_2020_CVPR} train split.
To compute ground truth detections, we compute the tight bounding box around pixel coordinates of the projected vertices for the objects and the hand.
For the hands, we post-process the detections to recover square boxes.
In Table.~\ref{tab:detect2d} we report the AP for hands and objects before and after tracking.
We observe that tracking~\cite{motpy} effectively allows us to recover valid detections. 
The use of tracking improves AP by 24\% and 12\% for hands and objects respectively.

\begin{table}[b]
\centering
\begin{tabular}{l|c|c}
\toprule
    & Hand AP &  Object AP \\
\midrule
Without tracking  & 0.61 & 0.59 \\
With tracking & 0.85 & 0.71 \\
\bottomrule
\end{tabular}
\caption{\textbf{Detection accuracy:} Hand and object detection precision using the 100DOH detector on the HO-3D~\cite{Hampali_2020_CVPR} train split. Tracking allows to retrieve reliable detections both for hands and objects.}
\label{tab:detect2d}
\end{table}

\begin{figure} %
\begin{center}
       \includegraphics[width=0.99\linewidth,bb=0 0 10cm 22cm]{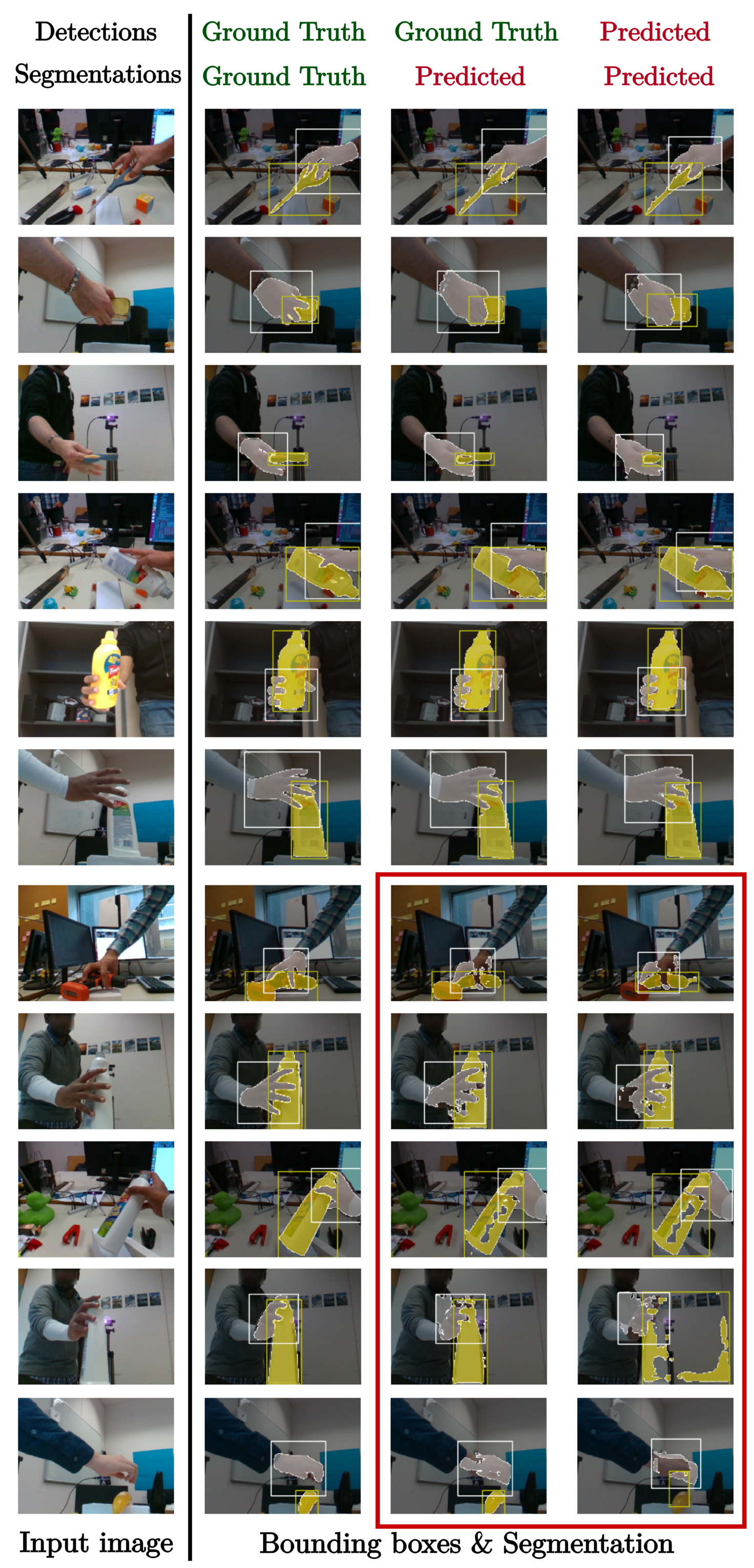}
  \caption{We qualitatively investigate the performance of the 2D detection and segmentation models on the HO-3D~\cite{Hampali_2020_CVPR} train split. Outlined in red, we display segmentation failures for hands or objects which are either due to imprecise detections or occur despite good detections.}
  \label{fig:segms2d}
\end{center}
\end{figure}

\subsection{Evaluation of hand and object segmentation}
\label{app:subsec:segm2d}

We evaluate the accuracy of our instance mask estimation procedure.
We report the IoU between the PointRend~\cite{kirillov2019pointrend} outputs and the ground truth segmentation masks, which we obtain by rendering the ground truth meshes.
As the hand ground truth meshes are unavailable for the test split, we report results on the HO-3D train split.
As described in the method section of the main paper, we select the mask produced for the most confident class prediction for the object and the Person class for the hand.
To investigate the segmentation error which is due to bounding box errors, we report IoU scores for the PointRend segmentation head both using the tracked and the ground truth detections in Table.~\ref{tab:segm2d}.
As expected, segmentation masks are more precise when using ground truth object boxes, especially for objects. Hence, our method would benefit from more robust and accurate detections.
We illustrate success and failures for 2D mask extraction in Figure~\ref{fig:segms2d}.

\begin{table}
\centering
\begin{tabular}{l|cc}
\toprule
Detections & Hand IoU &  Object IoU \\
\midrule
Ground Truth boxes & 0.63 & 0.74 \\
Predicted boxes~\cite{Shan20} & 0.61 & 0.66 \\
\bottomrule
\end{tabular}
\caption{\textbf{Segmentation accuracy:} Hand and object segmentation IoU for the PointRend mask estimates on the HO-3D dataset~\cite{Hampali_2020_CVPR}.}
\label{tab:segm2d}
\end{table}

\section{Implementation details.}
\label{app:sec:implem-details}

We describe details of our fitting implementation.
Our code and manually collected annotations will be made available upon publication.

\vspace{-.3cm}
\paragraph{Optimization procedure.}
Two of our error terms, the collision and fine-grained interactions act locally and require a reasonable initial configuration to provide meaningful gradients.
We therefore obtain the final hand-object reconstructions by performing two consecutive steps. We first recover a coarse pose for the hand and the object using the data and coarse interaction and regularization terms ($\mathcal{L}_{obj}, \mathcal{L}_{v2d}, \mathcal{L}_{pca}, \mathcal{L}_{scale}, \mathcal{L}_{smooth}, \mathcal{L}_{centroid}$). We then refine the optimized parameters using the full set of error terms.
Both steps are run for 200 optimization steps using the Adam~\cite{kingma:adam} optimizer.
We use different learning rates for various optimized parameters in order to distribute the updates between articulated and rigid pose updates. We use a fixed learning rate of $0.1$ for the PCA components of the MANO~\cite{mano} articulated hand model and the continuous rotation parameters~\cite{Zhou_2019_CVPR}. For the translation and scale updates we use a learning rate of $0.01$.

\vspace{-.3cm}
\paragraph{Error terms weighting.}
We use the same weights to balance the various error terms across all reconstructed datasets.
We empirically fix the weights to the following values.

\noindent Data weights: $\lambda_{obj}: 1$, $\lambda_{v2d}: 50$

\noindent Regularization weights: $\lambda_{pca}: 0.004 $, $\lambda_{scale}: 0.001 $, $\lambda_{smooth}: 2000 $

\noindent Interaction weights: $\lambda_{centroid}: 1 $, $\lambda_{local}: 1 $, $\lambda_{col}: 0.001$

\vspace{-.3cm}
\paragraph{Mesh preprocessing.} In order to reason about collisions and compute penetration depths, we preprocess the meshes to make them watertight. We further uniformly downscale the mesh to speed-up the rendering which is required to compute the object silhouette loss $\mathcal{L}_{obj}$.
We compute watertight meshes for all ShapeNet models~\cite{ShapeNet} using ManifoldPlus~\cite{huang2020manifoldplus}.
We rely on ACVD~\cite{ValetteCP08_acvd} to uniformly resample the object meshes, and set the target number of vertices to $1000$ which we observe empirically to provide a satisfactory trade-off mesh approximation and rendering speed.

\vspace{-.3cm}
\paragraph{Collision loss.}
Unlike PHOSA~\cite{phosa} which penalizes collisions using a \textit{local} loss~\cite{Ballan_2012_ECCV} which acts at the mesh surfaces, we use an implementation which relies on Signed Distance Fields (SDF)~\cite{Jiang_2020_CVPR}. We follow the original implementation, using a grid size of 32 for the signed distance field.

\vspace{-.3cm}
\paragraph{Local interaction heuristic.} We use the loss term introduced by ~\cite{hasson19_obman}, which showed improved qualitative results when used in the context of fully-supervised learning.
We accelerate computations by relying on the SDF computation from~\cite{Jiang_2020_CVPR} to perform collision checking.
We use the same parameters as in the original implementation for the attractive and repulsive terms, $1\textrm{cm}$ and $2\textrm{cm}$ respectively.

\vspace{-.3cm}
\paragraph{Contact accuracy.} We measure contact following~\cite{GraspingField:3DV:2020}, where contact is defined as negative hand vertex values in the object SDF.

\vspace{-.3cm}
\paragraph{Runtime.}
As we rely on learnt models for detection, segmentation and initial hand pose estimation, pre-processing 10 frames of dimension 640*480 takes less than a second on a Tesla V100 Nvidia GPU.
The initial object fitting stage for 50 random object pose initialization  takes around 10 second per frame.
The joint fitting runs at 14 iterations per iteration in absence of collision and contact terms.
With the fine-grained interaction constraints, the iteration time increases up to 4 iterations per second.
In total, complete fitting of a 10 frame video clip takes between 2 and 3 minutes.

\end{document}